\documentclass{preprint}

\usepackage{amssymb}
\usepackage{booktabs}
\usepackage{float}
\usepackage{multirow}

\title{ContextMaster: Interactive Multi-Shot Video Creation via Fixed-Budget Sparse Context Routing}

\author[1,\star]{Xu Guo}
\author[2,\star]{Zhengxuan Wei}
\author[3,\star,\S]{Xinghui Li}
\author[4]{Hanzhuo Huang}
\author[5]{Xinyu Liu}
\author[1]{Xiangyang Luo}
\author[3]{Min Wei}
\author[3]{Yiran Zhu}
\author[3]{Qiulin Wang}
\author[3]{Yulong Xu}
\author[3]{Xintao Wang}
\author[3]{Pengfei Wan}
\author[2]{Qi Fan}
\author[1,\S]{Xiangwang Hou}

\affiliation[1]{Tsinghua University}
\affiliation[2]{Nanjing University}
\affiliation[3]{Kling Team, Kuaishou Technology}
\affiliation[4]{ShanghaiTech University}
\affiliation[5]{The Hong Kong University of Science and Technology}

\contribution[\star]{Equal contribution}
\contribution[\S]{Corresponding author}

\abstract{%
Recent video models increasingly support generation, reference conditioning,
and editing within a single model, yet typically expose them as separate
operations over fixed inputs.
Practical creation unfolds across multiple shots, requiring one model to
generate from text, follow a reference, or edit source footage while maintaining
shared history. We formalize this setting as \emph{interactive multi-shot video
creation} (IMVC) and introduce ContextMaster, a unified model with a role-aware
context representation for these operations. An interactive model must retain
access to an expanding history without allowing the context read cost at each
denoising step to grow. ContextMaster combines reusable clean context states
with fixed budget sparse context routing and uses ConstraintSink to keep task
constraints visible. To address the dual challenges of sparse context access
and inference with few denoising steps, we propose a two-stage privileged
context distillation framework, which transfers full context behavior
from a dense teacher through consistency distillation and then refines
deployment rollouts with distribution matching. Experiments on the three
primitive tasks demonstrate improved task fulfillment and consistency across
shots over specialized baselines. User studies further validate flexibly
composed workflows, while the model reaches 16~FPS on a single GPU. Project Page: \url{https://guoxu1233.github.io/ContextMaster/} 
}

\begin{document}
\maketitle

\begin{figure*}[t]
    \centering
    \includegraphics[width=\textwidth]{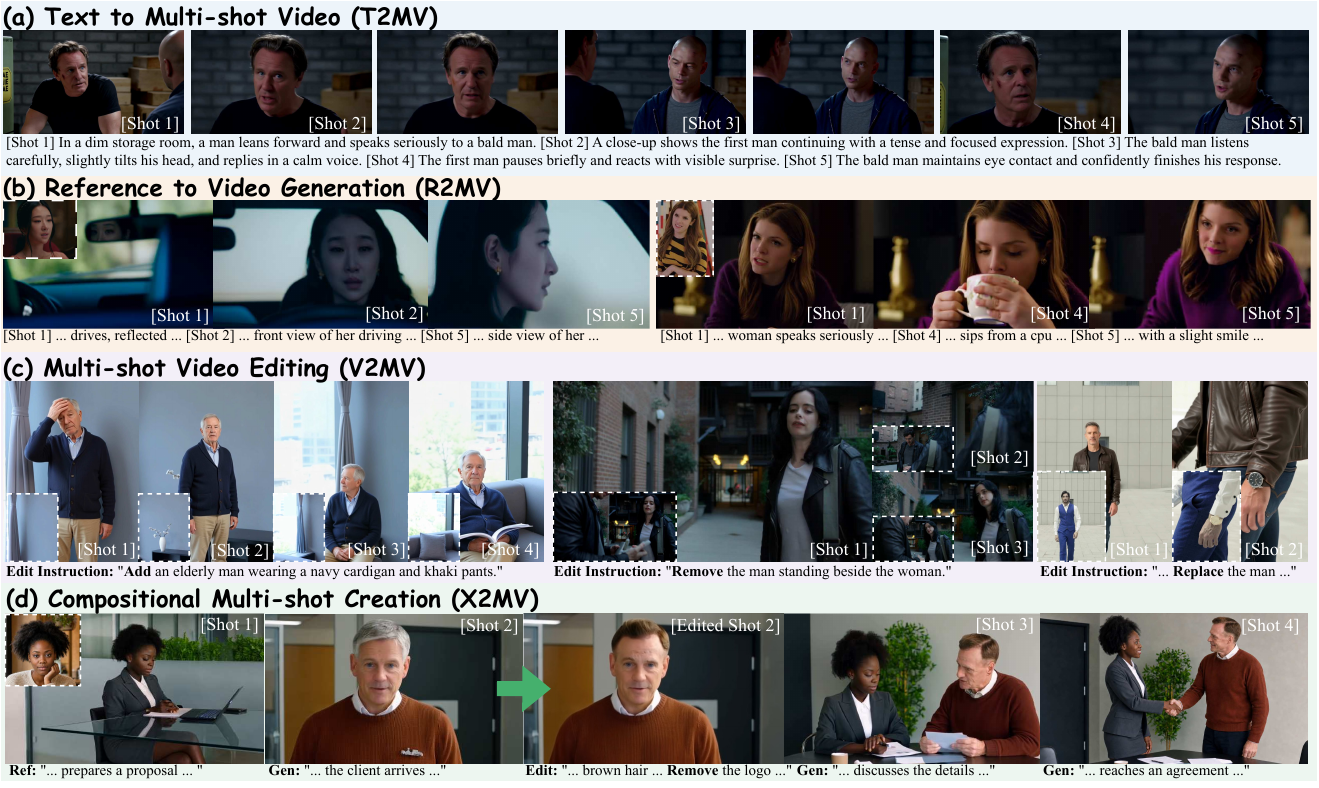}
    \caption{ContextMaster supports text to multi-shot video (T2MV), reference
    to multi-shot video (R2MV), multi-shot video editing (V2MV), and
    compositional multi-shot creation (X2MV), while achieving 16~FPS on a
    single GPU.}
    \label{fig:teaser}
\end{figure*}

\section{Introduction}

Recent proprietary and open video models increasingly integrate
text-driven generation, multimodal reference conditioning, and editing within
unified creation systems
\citep{seedance2026,klingteam2025omni,jiang2025vace,ye2026unic,
guo2026dreamidomni}.  These systems substantially broaden what can be created
in one model, but their capabilities are typically invoked as separate
operations over fixed inputs rather than composed over an evolving creation
state.  In this work, we formalize \emph{interactive multi-shot video creation}
(IMVC), a stateful multi-turn setting in which a single model can generate a
new shot, incorporate a visual reference, or edit an existing shot at each
turn, and then update the shared visual history with the accepted result.
These primitives can be composed in arbitrary order, supporting practical
workflows that interleave generation, editing, and continuation.

IMVC first requires a single model to process heterogeneous visual context.
A reference image specifies appearance, historical shots carry persistent
identity and narrative state, a source video provides frame-aligned editing
structure, and the noisy target represents content yet to be synthesized.
Concatenating them under one temporal coordinate system obscures their distinct
roles and conflates positional semantics across heterogeneous context streams.
We therefore introduce role-aware rotary coordinates that combine native frame
coordinates with role-specific and shot-specific phase offsets, allowing the
model to distinguish references, historical shots, sources, and
targets within a sequence.

Meanwhile, the stored context grows after every interaction round, causing
dense context reads repeated throughout denoising to increase with the session
length.  Interactive creation instead requires predictable denoising cost.
Streaming systems control this growth using sparsely sampled historical frames,
short rolling windows, or fixed sink frames
\citep{luo2026shotstream,yang2025longlive}.  These choices bound computation but
trade away historical coverage.  Recent retrieval methods recover distant
evidence through camera- or content-aware memory
\citep{yu2025cam,meng2026causalcine}, yet they are not designed for
heterogeneous reference conditions or frame-aligned editing constraints.  IMVC
requires bounded active context access while retaining the full
history as a candidate pool and keeping explicit reference and source
conditions visible.  We address these requirements with \emph{cacheable
fixed-budget context}.  An asymmetric
clean-context topology makes the observed context independent of target noise,
so its keys and values are prefilled once per round and reused during
denoising.  Block-sparse attention then limits each target query to a fixed
active-read budget over the stored context.  Within the same budget, a
\emph{ConstraintSink} reserves the provided reference and temporally aligned
source blocks, while the remaining capacity retrieves content-relevant source
and history blocks.

Cacheable fixed-budget context keeps active-read cost bounded as the stored
history grows, but each target block denoises from only a routed subset of the
stored evidence.  Reliable context selection is therefore critical.
Interactive latency further requires few-step sampling, leaving fewer updates
to compensate for imperfect sparse context use.  The deployment model must
thus learn both where to read and how to approximate dense full-context
behavior within a shortened trajectory.  We address this coupled problem with
\emph{privileged context distillation}.  A dense full-context teacher first
supervises the sparse deployment student through consistency distillation,
transferring full-context denoising behavior under sparse access while
establishing a stable few-step flow map.  The resulting student initializes a
subsequent distribution matching
refinement \citep{gu2025blade,yin2024dmd}, which trains on deployment matched
rollouts to recover perceptual detail and reduce errors accumulated in
generated history.

We quantitatively evaluate the three primitive capabilities on text to
multi-shot video (T2MV), reference to multi-shot video (R2MV), and multi-shot
video editing (V2MV).  We further validate compositional multi-shot creation
(X2MV), which combines these primitives over an evolving visual history.
ContextMaster improves task fulfillment and cross-shot consistency over
task-specific baselines while reaching 16 FPS on a single GPU.

Our main contributions are:
\begin{itemize}
    \item \textbf{Interactive multi-shot video creation.}
    We formulate IMVC as the stateful composition of generation,
    reference-guided generation, and editing over an evolving visual history,
    and introduce role-aware rotary coordinates to unify their heterogeneous
    context while preserving multi-shot alignment.
    \item \textbf{Cacheable fixed-budget context.}
    We combine reusable clean-context states, query-dependent block-sparse
    reads, and ConstraintSink to bound active context access without discarding
    explicit reference and source correspondences.
    \item \textbf{Privileged context distillation.}
    We use a dense full-context teacher to initialize a sparse few-step student
    through consistency distillation, followed by distribution matching
    refinement on deployment-matched rollouts.
\end{itemize}
\section{Related Work}

\paragraph{Unified and interactive video creation.}
Holistic multi-shot methods synthesize an entire planned sequence with
bidirectional access across shots
\citep{meng2025holocine,wang2026multishotmaster,huang2026poco}, whereas
streaming methods expose a next-shot interface conditioned on historical
memory \citep{luo2026shotstream,huang2026unityshots,meng2026causalcine,
an2026onestory}.
In parallel, unified models and editors support reference-conditioned
generation and diverse video manipulations within shared architectures
\citep{jiang2025vace,ye2026unic,shao2026liveditor,wang2026liveedit,
guo2026dreamidv,wei2026dreamvideo,liu2025revise,liu2026rebind,
zhang2026omnitransfer,mou2025instructx,luo2026cointeract}.
These paradigms generally expose generation, reference conditioning, and
editing as separate interfaces.  IMVC instead treats them as stateful
operations that can be composed over an evolving visual history.

\paragraph{Heterogeneous and growing visual context.}
Prior work uses discontinuous, phase-shifted, or group-specific rotary
coordinates to distinguish shot identities and in-context conditions
\citep{luo2026shotstream,wang2026multishotmaster,huang2026poco,
shao2026liveditor}.  Our role-aware coordinates extend this direction to a
shared interface containing reference, history, source, and target streams.
For growing history, existing systems use sparse frame sampling, rolling
windows, fixed sinks, content-aware retrieval, or learned memory mechanisms
\citep{luo2026shotstream,yang2025longlive,yu2025cam,meng2026causalcine,
wei2026geometryawareimplicitmemoryvideo}.
Our fixed-budget formulation retains the stored history as routing candidates
while bounding each target query's active reads, with explicit conditions and
dynamic retrieval sharing the same capacity.

\paragraph{Sparse and few-step video diffusion.}
Sparse video attention exploits spatiotemporal patterns or retrieves
content-relevant blocks, and BLADE combines adaptive block sparsity with
sparsity-aware step distillation
\citep{xi2025sparsevideogen,shao2026liveditor,gu2025blade}.  Separately,
consistency and distribution matching objectives enable few-step generation
\citep{song2023consistency,wang2024pcm,yin2024dmd,yin2025causvid}; ShotStream
further uses self-forcing to reduce causal rollout errors
\citep{huang2025selfforcing,luo2026shotstream}.
Our setting differs in that the teacher reads complete clean context, whereas
the deployment student uses a fixed budget.  We transfer the teacher's
denoising behavior through consistency distillation and then refine sparse
few-step rollouts through distribution matching.
\section{Methodology}

Our framework operates on an evolving shot history through a shared
generation-and-editing interface.  It has three components.  First, a
role-aware representation places reference, history, source, and target
latents in one model without conflating their temporal semantics.  Second, a
cacheable fixed-budget operator encodes the context once per interaction
round and bounds every target query's active reads.  Third, privileged context
distillation transfers dense full-context behavior to the sparse few-step
student used at deployment.

\begin{figure*}[t]
    \centering
    \includegraphics[width=\textwidth]{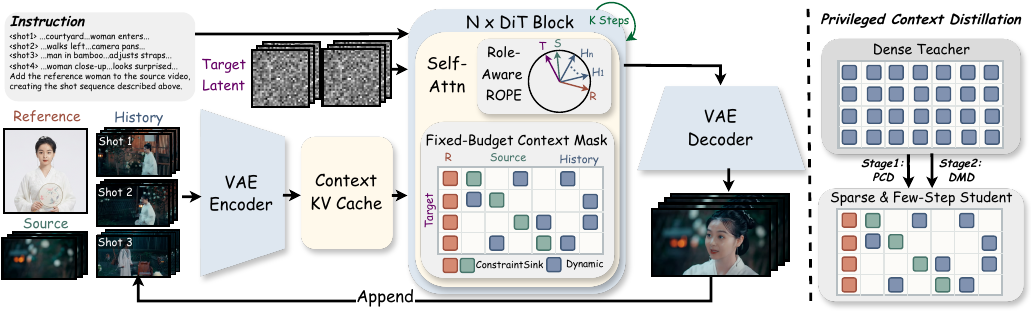}
    \caption{Overview of our unified IMVC framework.  Reference, history, and
    optional source latents form clean context whose keys and values are cached
    across denoising steps.  Role-aware RoPE distinguishes their temporal
    roles, while fixed-budget attention retains mandatory correspondences and
    dynamically retrieves the remaining context.  Only the target latents are
    iteratively updated and decoded.  Each accepted output is appended to the
    history as clean context for subsequent rounds.  The right panel summarizes
    the two-stage PCD and DMD training strategy.}
    \label{fig:framework}
\end{figure*}

\subsection{IMVC Formulation and Role-Aware Context}

At interaction round $r$, the model receives an instruction $u_r$, an optional
reference image $R_r$, an optional source video $S_r$, and the shot history
$H^{(r-1)}$.  It produces
\begin{equation}
    \widehat{V}_r =
    {\cal G}_{\theta}\!\left(
        u_r,R_r,S_r,H^{(r-1)}
    \right).
    \label{eq:task}
\end{equation}
Without $S_r$, $\widehat{V}_r$ is a new shot, optionally guided by $R_r$; with
$S_r$, it is an edited version of the source.  We distinguish persistent
history from request-local visual conditions: $H^{(r-1)}$ contains only
finalized shots, while neither $R_r$ nor $S_r$ is stored in it.  When
$\widehat{V}_r$ is accepted, it is appended to the history and re-encoded as
clean context in the next round.  A
rejected candidate may instead serve as $S_{r+1}$ for further editing without
first entering the history.  Under this common rule, T2MV and R2MV commit
accepted generations, V2MV commits accepted edits, and X2MV composes these
request types over successive rounds.

These operations expose visual inputs with different temporal semantics:
references specify appearance, history carries identity and narrative state,
and source videos provide frame-aligned editing structure.  We encode the
available inputs with the video VAE, concatenate their latents by role, and
apply patch embedding ${\cal P}$.  We omit the round index $r$ from the
architectural notation below:
\begin{equation}
    [\,C^0;X^0\,]
    =
    {\cal P}\!\left(
        [\,Z^{\mathrm{ref}};
           Z^{\mathrm{hist}};
           Z^{\mathrm{src}};
           Z_{\sigma}^{\mathrm{tar}}\,]
    \right),
    \label{eq:layout}
\end{equation}
where $C^0$ contains the observed reference, history, and source tokens, and
$X^0$ contains the target tokens noised at diffusion time $\sigma$; absent
roles occupy no tokens.

\paragraph{Role-aware RoPE.}
Motivated by the discontinuous RoPE in
ShotStream~\cite{luo2026shotstream}, we design a phase offset that identifies the
role of each visual segment in the concatenated sequence.  Each token $i$ is
assigned a frame coordinate $t_i$ and a role--shot code $c_i$.  For temporal
frequency index $\nu$, its rotary phase is
\begin{equation}
    \phi_{i,\nu} = \omega_\nu t_i + \alpha c_i ,
    \label{eq:role-rope}
\end{equation}
where $\omega_\nu$ is the standard RoPE frequency~\cite{su2024roformer} and $\alpha$ controls the
phase offset.  For a request with $n$ historical shots, we set $c_i=-1$ for
the reference, $c_i=j$ for the $j$-th historical shot,
$c_i=n+\frac{1}{2}$ for the source video, and $c_i=n+1$ for the target.
The reference uses $t_i=0$,
whereas video tokens use their frame coordinates.  In editing, source and
target frames at time $f$ share $t_i=f$: the temporal term preserves their
alignment, while the offset tells the model whether a token belongs to the
reference, history, source, or target.

\subsection{Cacheable Fixed-Budget Context}

\paragraph{Cacheable sparse context attention.}
At each transformer layer, let $C$ and $X$ denote the context and target
states, and let $Q_U=UW_Q$, $K_U=UW_K$, and $V_U=UW_V$ for
$U\in\{C,X\}$.  After applying the role-aware RoPE in
Equation~(\ref{eq:role-rope}), we use asymmetric attention:
\begin{equation}
    \begin{array}{l}
    Y_C =
        \mathrm{Attn}(
            \widetilde Q_C,
            \widetilde K_C,
            V_C),\\[1mm]
    Y_X =
        \mathrm{Attn}\!\left(
            \widetilde Q_X,
            [\widetilde K_C;\widetilde K_X],
            [V_C;V_X]
        \right).
    \end{array}
    \label{eq:clean-context}
\end{equation}
As shown in Figure~\ref{fig:framework}, the context branch forms a reusable KV
cache.  It neither reads the noisy target nor performs text cross-attention.
We apply zero-timestep modulation to $C$ and diffusion-timestep modulation to
$X$, making $(\widetilde K_C,V_C)$ independent of both the denoising step and
CFG branch.  They are prefilled once per round and reused during sampling.

Caching removes repeated context encoding, but a dense target-to-context read
still grows with the shot history.  We therefore sparsify only this attention
band, while keeping context self-attention and target self-attention dense.
We partition both branches into hardware-aligned blocks of $m$ tokens.  For
target block $q$ and context block $k$, their head-shared relevance score is
\begin{equation}
    s_{qk}
    =
    \frac{1}{N_{\mathrm{head}}}
    \sum_{h=1}^{N_{\mathrm{head}}}
    \left\langle
        \overline{\mathbf q}_{q,h},
        \overline{\mathbf k}_{k,h}
    \right\rangle ,
    \label{eq:block-score}
\end{equation}
where $N_{\mathrm{head}}$ is the number of heads and
$\overline{\mathbf q}_{q,h}$ and $\overline{\mathbf k}_{k,h}$ are normalized
mean-pooled block summaries.  This score retrieves
content-dependent evidence.  Explicit reference and source--target
correspondences, however, should not rely only on similarity ranking; we
reserve budget for them and route the remaining blocks adaptively.

\paragraph{ConstraintSink.}
Explicit reference and source constraints should remain visible regardless of
content-relevance scores.  We therefore collect the provided reference and
exactly aligned source blocks into one mandatory set.  Let
${\cal C}_{\mathrm{ref}}$ and ${\cal C}_{\mathrm{src}}$ denote the reference
and source block sets, and let $t_q$ and $t_k$ be the center-frame coordinates
of target block $q$ and source block $k$.  We define
\begin{equation}
    {\cal A}_{\mathrm{sink}}(q)
    =
    {\cal C}_{\mathrm{ref}}
    \cup
    \left\{
        k\in{\cal C}_{\mathrm{src}}:
        t_k=t_q
    \right\},
    \label{eq:constraint-sink}
\end{equation}
The first term makes the reference visible to every target block, while the
second forms an exactly aligned source--target band that preserves motion and
structure during editing.  Missing
roles contribute an empty set.  All retained blocks consume the same context
budget.

\paragraph{Budgeted dynamic routing.}
We use an absolute budget of $B$ context blocks, rather than a fraction of the
growing context.  By construction, $|{\cal A}_{\mathrm{sink}}(q)|\leq B$; we
denote the residual capacity by
$B_q^{\mathrm{rem}}=B-|{\cal A}_{\mathrm{sink}}(q)|$.

We route this residual capacity separately within source and history.  For an
editing request with both source and history present, up to a fixed source
quota is assigned to content-dependent retrieval from source blocks outside
the ConstraintSink, and history receives the remaining capacity.  Without a
source video, the entire residual budget is assigned to history.  The
allocation is work-conserving:
if either role is absent or has fewer candidates than its allocation, the
unused capacity is transferred to the other role.  Denoting the resulting
cardinalities by $K_{\mathrm{src}}(q)$ and $K_{\mathrm{hist}}(q)$, the active
context is
\begin{equation}
    \begin{array}{rl}
    {\cal M}_B(q)={}&{\cal A}_{\mathrm{sink}}(q)\\
    &{}\cup
    \mathrm{TopK}_{K_{\mathrm{src}}(q)}
    \left({\cal C}_{\mathrm{src}}
    \setminus{\cal A}_{\mathrm{sink}}(q);s_{q\cdot}\right)\\
    &{}\cup
    \mathrm{TopK}_{K_{\mathrm{hist}}(q)}
    \left({\cal C}_{\mathrm{hist}};s_{q\cdot}\right).
    \end{array}
    \label{eq:budget-route}
\end{equation}
Here ${\cal C}_{\mathrm{src}}$ and ${\cal C}_{\mathrm{hist}}$ are the source
and history block sets, respectively, and
$K_{\mathrm{src}}(q)+K_{\mathrm{hist}}(q)\leq B_q^{\mathrm{rem}}$, with
equality whenever enough candidates exist.  Separating the two competitions
prevents the strong local similarity of source frames from suppressing
long-range history,
without requiring cross-role score calibration.  Every target block attends
to ${\cal M}_B(q)$ and to all target tokens.  Mandatory and dynamically routed
blocks share the same budget, so $|{\cal M}_B(q)|\leq B$ for every query.

Each target query activates at most $B$ blocks, or $Bm$ context tokens.  Let
$D$ denote the hidden width.  For $N_C$ context tokens and $N_X$ target tokens,
dense target-to-context attention costs ${\cal O}(N_X N_C D)$ per layer,
whereas sparse reads cost ${\cal O}(N_X B m D)$.  The active read cost is
therefore independent of history length.  Routing uses inexpensive block
summaries.  This guarantee targets the latency-critical reads repeated at
every denoising step.  Context states are prefilled once per round and reused
throughout the trajectory.

\subsection{Privileged Context Distillation}

Aggressive context sparsity and few-step sampling introduce complementary
errors: the student observes only a subset of the teacher's context and must
approximate a long flow trajectory with few evaluations.  Direct sparse flow
matching provides a per-timestep velocity target, but neither transfers the
dense teacher's full-context behavior nor enforces consistency across noise
levels.  We therefore use a frozen dense teacher ${\cal T}$ with unrestricted
context as privileged supervision for a sparse student ${\cal S}_\theta$
constrained by Equation~(\ref{eq:budget-route}).  Training proceeds in two
stages, as summarized in  Figure~\ref{fig:framework}.

\paragraph{Privileged consistency distillation (PCD).}
Let $x_0$ be a clean target latent and $\epsilon$ Gaussian noise.  We sample
adjacent noise levels $\sigma>\sigma'$ from a fine discretization of the same
shifted flow schedule~\cite{lipman2023flowmatching} used at deployment and construct
\begin{equation}
    x_\sigma = (1-\sigma)x_0 + \sigma\epsilon .
    \label{eq:flow-noise}
\end{equation}
The dense teacher evaluates conditional and unconditional full-context
velocities, which are combined by classifier-free guidance (CFG) \cite{ho2022classifier} into
$v_{\cal T}^{\mathrm{cfg}}$.  One online Euler step gives
\begin{equation}
    x_{\sigma'}^{\cal T}
    =
    x_\sigma
    +
    (\sigma'-\sigma)
    v_{\cal T}^{\mathrm{cfg}}
    (x_\sigma,C_{\mathrm{full}},u).
    \label{eq:teacher-step}
\end{equation}
The budgeted student and its exponential-moving-average target
${\cal S}_{\bar\theta}$ predict clean samples at the two endpoints:
\begin{equation}
    \begin{array}{l}
    \widehat{x}_{0}^{\,\theta}
    =
    x_\sigma
    -
    \sigma
    v_{{\cal S}_\theta}(x_\sigma,C_B,u),\\[2mm]
    \widehat{x}_{0}^{\,\bar\theta}
    =
    x_{\sigma'}^{\cal T}
    -
    \sigma'
    v_{{\cal S}_{\bar\theta}}
    (x_{\sigma'}^{\cal T},C_B,u).
    \end{array}
    \label{eq:cd-predictions}
\end{equation}
Here $C_{\mathrm{full}}$ and $C_B$ contain the same stored conditions, but only
the former permits unrestricted reads.  We minimize
\begin{equation}
    {\cal L}_{\mathrm{CD}}
    =
    \left\|
        \widehat{x}_{0}^{\,\theta}
        -
        \mathrm{sg}
        \left[
            \widehat{x}_{0}^{\,\bar\theta}
        \right]
    \right\|_2^2 ,
    \label{eq:cd-loss}
\end{equation}
where $\mathrm{sg}$ stops gradients.  The target parameters are updated as
$\bar\theta\leftarrow\mu\bar\theta+(1-\mu)\theta$.  Matching the two endpoints
makes the sparse student's clean prediction invariant to a full-context
teacher transition, transferring both privileged context and cross-timestep
consistency.  Because $x_{\sigma'}^{\cal T}$ is produced by a CFG-guided
teacher while both student evaluations use only conditional text, guidance is
absorbed into the student and requires no unconditional forward pass at
inference.

\paragraph{Distribution refinement.}
Consistency distillation establishes the few-step flow map but can smooth out
fine appearance and editing details.  We therefore refine the student with
distribution matching distillation (DMD)~\cite{yin2024dmd}.  Given a generated
clean prediction $\widetilde{x}_0$, we sample
$\epsilon_\gamma\sim{\cal N}(0,I)$ and form
$y_\gamma=(1-\gamma)\widetilde{x}_0+\gamma\epsilon_\gamma$.  A frozen dense
real-score model with full context and CFG and a trainable fake-score model
with budgeted context define
\begin{equation}
    \begin{array}{l}
    \Delta s_\gamma =
    s_{\mathrm{fake}}(y_\gamma,\gamma,C_B,u)
    -
    s_{\mathrm{real}}^{\mathrm{cfg}}
    (y_\gamma,\gamma,C_{\mathrm{full}},u),\\[1mm]
    \displaystyle
    \nabla_\theta{\cal L}_{\mathrm{DMD}}
    =
    \mathop{\mathrm{E}}_{\gamma,\epsilon_\gamma}
    \left[
        w(\gamma)\Delta s_\gamma
        \frac{\partial\widetilde{x}_0}{\partial\theta}
    \right],
    \end{array}
    \label{eq:dmd-gradient}
\end{equation}
where $w(\gamma)$ is a timestep-dependent weight.  Under our flow
parameterization, $\Delta s_\gamma$ is proportional to the difference between
the fake- and real-score clean predictions.  We use its detached,
per-sample-normalized form and the standard stop-gradient surrogate to pass
this direction to the student; the fake score is trained on current student
samples.

For each update, we uniformly sample an exit index
$e\in\{0,\ldots,N_{\mathrm{step}}-1\}$ and run the fixed-budget student along
the deployment noise schedule up to $e$.  Earlier transitions are detached,
and gradients pass only through the clean prediction at the sampled exit.
This exposes every deployment step to distribution-level supervision while
preserving the inference-time sparse topology and context budget.  Moreover,
the shot history is autoregressively generated by the current student through
complete $N_{\mathrm{step}}$-step rollouts.  Each generated shot is decoded,
resampled, and re-encoded by the same context pipeline used at inference.
Distribution refinement therefore accounts for both few-step sampling errors
and errors accumulated in the student's own history.

\subsection{Interactive Inference}

The interface in Equation~(\ref{eq:task}) can be operated directly by a user or
invoked by an optional tool-augmented Director.  The latter converts interactive
creation into an automated workflow: it plans scripts and shot sequences,
retrieves reference images, and calls shot-boundary and temporal-localization
tools to identify the source segments to edit.  It then executes the resulting
shot specifications $(u_r,R_r,S_r)$ through the same video model.  The planning
and tool calls remain external to the video model and do not alter its
latency-critical denoising path.  Further implementation details are provided
in Appendix~\ref{app:interactive-director}.
\section{Experiments}

\begin{figure*}
    \centering
    \includegraphics[width=1\linewidth]{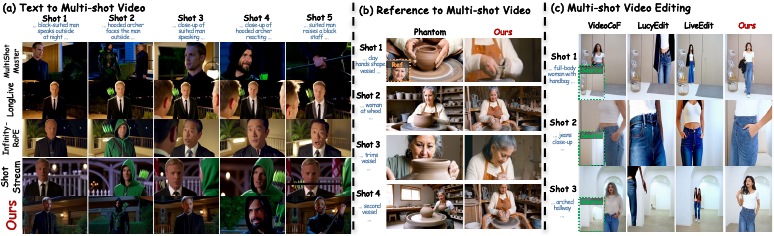}
    \caption{Qualitative comparison across text to multi-shot generation
    (T2MV), reference to multi-shot generation (R2MV), and multi-shot video editing
    (V2MV).  Frames are sampled from successive shots.  ContextMaster retains
    recurring characters under viewpoint changes, follows the reference
    identity throughout a multi-shot action, and applies edits consistently
    while preserving source structure. Please zoom in for more details.}
    \label{fig:qualitative-comparison}
\end{figure*}

\subsection{Setup}

\paragraph{Benchmarks.}
We evaluate all IMVC task modes with task-specific test sets.  The
T2MV set contains 100 multi-shot prompts with recurring
characters, scene changes, and alternating viewpoints.  The
R2MV set contains 50 reference-conditioned
cases, each specifying five successive shots.  The
V2MV editing set contains 50 multi-shot source video cases spanning subject addition, removal, and replacement. For X2MV, we collect 20 interactive trajectories, each combining at least two operations over the evolving history, and evaluate the resulting sequences in the user study. All methods
receive the same shot-level instructions and the same reference or source
inputs when required.

\paragraph{Implementation details.}
Our model is initialized from Wan2.1-T2V-1.3B~\citep{wan2025}, to generate 832 × 480 video clips.  We generate 81-frame each shot and uniformly sample each accepted shot at 1~FPS to build an append clean-context memory.  The deployed
student uses four denoising steps, baked-in guidance, and a fixed active-read
budget of six frame equivalents (6~FE).  At inference, following
LIVEditor~\citep{shao2026liveditor}, we implement the routed sparse-attention
forward pass with a block-wise zeroth-order Taylor kernel in
TileLang~\citep{wang2025tilelang}. The same sparse operator and context
budget are used throughout privileged consistency distillation, distribution
refinement, and inference.
The model is trained on an internal dataset of 1M multi-shot videos, and public datasets for reference and editing.
Further details on training and data construction are provided in the
supplementary material.

\paragraph{Evaluation metrics.}
We evaluate video quality using text alignment (TA) from
ViCLIP~\citep{wang2023internvid} and aesthetic quality (AQ), motion smoothness
(MS), and dynamic degree (DD) from VBench~\citep{huang2024vbench}.  For
intra-shot consistency, subject and background similarity are computed with
DINO~\citep{caron2021dino} and CLIP~\citep{radford2021clip} image features,
respectively.  Following HoloCine~\citep{meng2025holocine}, Inter-Shot groups
shots containing the same character and compares their ViCLIP features.
Gemini~2.5~Pro~\citep{geminiteam2024} rates task fulfillment (TF) on a 1--5
scale using task-specific criteria.  Following
ShotStream~\citep{luo2026shotstream}, We also evaluate inference efficiency of all methods
using a single H200 GPU under
matched resolution, frame count.

\subsection{Comparison}

\begin{table*}[t]
    \centering
    \small
    \setlength{\tabcolsep}{2.5pt}
    \begin{tabular*}{\textwidth}{@{\extracolsep{\fill}}llccccccccc@{}}
        \toprule
        Task
        & Method
        & \multicolumn{4}{c}{Video Quality}
        & \multicolumn{3}{c}{Consistency}
        & Fulfillment
        & Efficiency \\
        \cmidrule(lr){3-6}
        \cmidrule(lr){7-9}
        \cmidrule(lr){10-10}
        \cmidrule(lr){11-11}
        &
        & TA$\uparrow$
        & AQ$\uparrow$
        & MS$\uparrow$
        & DD$\uparrow$
        & Intra-Sub.$\uparrow$
        & Intra-Bg.$\uparrow$
        & Inter-Shot$\uparrow$
        & TF$\uparrow$
        & FPS$\uparrow$ \\
        \midrule
        \multirow{5}{*}{T2MV}
        & MultiShotMaster
            & 0.179
            & 0.513
            & 0.973
            & \underline{0.617}
            & 0.819
            & 0.872
            & 0.739
            & 3.94
            & 0.19 \\
        & LongLive
            & 0.163
            & 0.529
            & 0.992
            & 0.223
            & 0.949
            & 0.942
            & 0.795
            & 3.54
            & \underline{16.55} \\
        & Infinity-RoPE
            & \underline{0.191}
            & \underline{0.560}
            & 0.988
            & 0.569
            & 0.827
            & 0.873
            & 0.711
            & 3.86
            & 16.37 \\
        & ShotStream
            & 0.165
            & 0.535
            & \textbf{0.996}
            & 0.346
            & \textbf{0.986}
            & \textbf{0.979}
            & \underline{0.808}
            & \underline{4.03}
            & 15.95 \\
        & \textbf{Ours}
            & \textbf{0.205}
            & \textbf{0.565}
            & \underline{0.994}
            & \textbf{0.669}
            & \underline{0.980}
            & \underline{0.974}
            & \textbf{0.836}
            & \textbf{4.17}
            & \textbf{16.74} \\
        \midrule
        \multirow{2}{*}{R2MV}
        & Phantom
            & \underline{0.196}
            & \underline{0.558}
            & \underline{0.989}
            & \textbf{0.787}
            & \textbf{0.953}
            & \underline{0.951}
            & \underline{0.638}
            & \underline{3.68}
            & \underline{0.452} \\
        & \textbf{Ours}
            & \textbf{0.199}
            & \textbf{0.585}
            & \textbf{0.992}
            & \underline{0.689}
            & \underline{0.946}
            & \textbf{0.955}
            & \textbf{0.749}
            & \textbf{4.04}
            & \textbf{16.74} \\
        \midrule
        \multirow{5}{*}{V2MV}
        & VideoCoF
            & \underline{0.223}
            & \textbf{0.533}
            & 0.993
            & 0.494
            & 0.909
            & 0.939
            & \underline{0.714}
            & \underline{3.80}
            & 1.41 \\
        & LucyEdit
            & 0.171
            & 0.461
            & 0.992
            & \underline{0.516}
            & \underline{0.962}
            & 0.944
            & 0.623
            & 2.96
            & 1.05 \\
        & StreamEdit
            & 0.156
            & 0.475
            & 0.984
            & 0.483
            & 0.924
            & 0.928
            & 0.632
            & 3.10
            & 3.24 \\
        & LiveEdit
            & 0.197
            & 0.477
            & 0.991
            & 0.419
            & 0.961
            & \underline{0.954}
            & 0.638
            & 3.26
            & \underline{12.66} \\
        & \textbf{Ours}
            & \textbf{0.238}
            & \underline{0.520}
            & \textbf{0.994}
            & \textbf{0.548}
            & \textbf{0.969}
            & \textbf{0.972}
            & \textbf{0.751}
            & \textbf{4.20}
            & \textbf{16.74} \\
        \bottomrule
    \end{tabular*}
    \caption{Unified comparison on Text to Multi-shot Video (T2MV), Reference
    to Multi-shot Video (R2MV), and Multi-shot Video Editing (V2MV). TA, AQ,
    MS, DD, and TF denote text alignment, aesthetic quality, motion smoothness,
    dynamic degree, and task fulfillment, respectively. TF is rated by
    Gemini~2.5~Pro on a 1--5 scale. Best and second-best results are marked in
    bold and underlined within each task.  The reported 16.74~FPS is averaged over
five-shot runs across T2MV, R2MV, and V2MV. }
    \label{tab:unified-comparison}
\end{table*}

\paragraph{Comparison on T2MV.}
We compare with the holistic bidirectional model
MultiShotMaster~\citep{wang2026multishotmaster}, the real-time autoregressive
models LongLive~\citep{yang2025longlive} and
ShotStream~\citep{luo2026shotstream}, and Infinity-RoPE~\citep{yesiltepe2026infinity}.  As shown in
Table~\ref{tab:unified-comparison}, our model provides the strongest overall
performance, including an Inter-Shot gain from 0.808 to 0.836, while remaining
competitive on intra-shot consistency.  Figure~\ref{fig:qualitative-comparison}(a)
reveals the corresponding failure mode: competing methods substitute an
identity or lose one participant after viewpoint changes.  Our model keeps the
suited man and hooded archer visually distinct throughout the alternating
five-shot sequence.

\paragraph{Comparison on R2MV.}
We compare with Phantom~\citep{liu2025phantom}, a reference conditioned
subject-consistent generator, using the same
references and shot descriptions.  Our largest gains occur in Inter-Shot and
TF, while reaching 16.74~FPS; Phantom retains higher DD and slightly higher
intra-shot subject consistency.  In
Figure~\ref{fig:qualitative-comparison}(b), Phantom shows less stable reference
appearance across close-up and wider views.  Our model preserves the elderly
potter while following the progression from shaping and trimming the first
vessel to starting a second one.

\paragraph{Comparison on V2MV.}
We compare with the temporal-reasoning editor
VideoCoF~\citep{yang2026videocof}, the instruction editor
LucyEdit~\citep{decart2025lucyedit} and the real-time streaming editor
StreamEdit~\citep{jiao2026streamedit}, 
LiveEdit~\citep{wang2026liveedit}, using the same source shots and instructions.
Our model leads or ties on eight of the nine reported metrics, while VideoCoF retains a small AQ advantage.  As shown in
Figure~\ref{fig:qualitative-comparison}(c), competing editors produce incomplete
or view-dependent appearance changes.  Our method applies the requested edit
consistently across full-body, close-up, and wide shots while preserving the
source scene structure.

\subsection{User Study}

We invite 10 video professionals to evaluate three primitive tasks and X2MV in
a common user study.  Participants rate visual quality (VQ), instruction
following (IF), temporal consistency (TC), and cross shot consistency (CC) with
scores from 1 to 5.  We retain the strongest baseline for each primitive task.

\begin{table}[H]
    \centering
    \small
    \setlength{\tabcolsep}{2pt}
    \begin{tabular*}{\columnwidth}{@{\extracolsep{\fill}}llcccc@{}}
        \toprule
        Task & Method
        & VQ$\uparrow$
        & IF$\uparrow$
        & TC$\uparrow$
        & CC$\uparrow$ \\
        \midrule
        T2MV & ShotStream
            & 3.76 & 3.68
            & \textbf{4.14} & 3.74 \\
        & Ours
            & \textbf{3.96} & \textbf{3.91}
            & 4.10 & \textbf{4.04} \\
        \midrule
        R2MV & Phantom
            & 3.65 & 3.35
            & 3.94 & 3.42 \\
        & Ours
            & \textbf{4.00} & \textbf{3.86}
            & \textbf{4.11} & \textbf{3.99} \\
        \midrule
        V2MV & VideoCoF
            & \textbf{3.70} & 3.45
            & 3.89 & 3.52 \\
        & Ours
            & 3.61 & \textbf{4.00}
            & \textbf{4.12} & \textbf{3.98} \\
        \midrule
        X2MV & Ours
            & 3.77 & 3.84
            & 4.04 & 3.65 \\
        \bottomrule
    \end{tabular*}
    \caption{Unified user study across four IMVC settings.  Mean ratings are
    reported on a 1--5 scale.}
    \label{tab:user-study}
\end{table}

\paragraph{Results.}
As shown in Table~\ref{tab:user-study}, our model obtains the highest IF and CC
scores against the strongest baseline across the three primitive tasks.  On
X2MV, scores remain favorable across mixed creation trajectories.

\subsection{Ablation Studies}

\begin{table}[H]
    \centering
    \small
    \setlength{\tabcolsep}{3pt}
    \begin{tabular*}{\columnwidth}{@{\extracolsep{\fill}}lcccc@{}}
        \toprule
        \multicolumn{5}{c}{(a) Architecture} \\
        \cmidrule(lr){1-5}
        Variant & Inter-Shot$\uparrow$ & Gen.\ TF$\uparrow$ & Ref.\ TF$\uparrow$ & Edit TF$\uparrow$ \\
        \midrule
        w/o role-aware RoPE
            & \underline{0.748} & 3.98 & \underline{3.82} & \underline{3.93} \\
        w/o ConstraintSink
            & 0.742 & \underline{4.06} & 3.69 & 3.64 \\
        \textbf{Ours}
            & \textbf{0.779} & \textbf{4.17} & \textbf{4.04} & \textbf{4.20} \\
        \bottomrule
    \end{tabular*}

    \medskip

    \begin{tabular*}{\columnwidth}{@{\extracolsep{\fill}}lcccc@{}}
        \toprule
        \multicolumn{5}{c}{(b) Training Strategy (T2MV)} \\
        \cmidrule(lr){1-5}
        Variant & AQ$\uparrow$ & MS$\uparrow$ & Inter-Shot$\uparrow$ & TF$\uparrow$ \\
        \midrule
        Vanilla sparse
            & 0.536 & 0.983 & 0.754 & 3.92 \\
        PCD only
            & 0.548 & \underline{0.992} & \underline{0.821} & \underline{4.05} \\
        DMD only
            & \underline{0.561} & 0.988 & 0.785 & 3.91 \\
        \textbf{PCD$\rightarrow$DMD (Ours)}
            & \textbf{0.565} & \textbf{0.994} & \textbf{0.836} & \textbf{4.17} \\
        \bottomrule
    \end{tabular*}
    \caption{Ablation of architectural and training choices.}
    \label{tab:ablation}
\end{table}

\begin{figure}[H]
    \centering
    \includegraphics[width=1\linewidth]{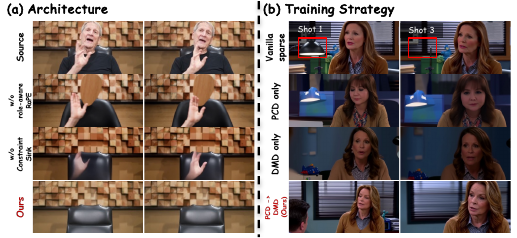}
    \caption{Qualitative ablations of architecture and training strategy.
    (a) Removing role-aware RoPE or ConstraintSink leaves visible source
    remnants after removal, whereas the full model restores the chair and
    background.  (b) PCD and DMD provide complementary benefits in cross-shot
    identity and visual detail. Please zoom in for more details.}
    \label{fig:ablation}
\end{figure}

\paragraph{Ablation on architecture.}
We keep the 6-FE budget fixed and compare two variants.  Without role-aware
RoPE, all visual tokens are directly concatenated under shared positional
coordinates.  Without ConstraintSink, the entire $B$-block budget is filled
solely by similarity-based Top-$k$ routing, with no mandatory reference or
aligned source blocks.  Inter-Shot in Table~\ref{tab:ablation}(a) is
macro-averaged across the three tasks.  The results show that role separation
benefits all task modes, whereas mandatory routing is particularly important
for reference-guided generation and editing.  Figure~\ref{fig:ablation}(a)
illustrates the latter failure: pure Top-$k$ routing leaves visible source
remnants, while the complete model removes the subject and reconstructs the
chair and background.

\paragraph{Ablation on training strategy.}
We conduct the training ablation on T2MV.  Vanilla sparse initializes the
student from the full-context model and directly trains it with sparse
attention.  PCD only performs privileged consistency distillation without the
subsequent DMD stage, whereas DMD only applies distribution matching directly
without PCD initialization.  As shown in Table~\ref{tab:ablation}(b), PCD
primarily improves Inter-Shot and TF, while DMD contributes more strongly to
appearance quality.  Their sequential combination performs best on every
metric.  Figure~\ref{fig:ablation}(b) shows the same complementarity: vanilla
sparse training produces local artifacts, and either single-stage variant
sacrifices appearance detail or cross-shot identity.  PCD followed by DMD
preserves the recurring woman while retaining sharper visual detail.

\section{Conclusion}

We introduced ContextMaster and formalized interactive multi-shot video
creation as the stateful composition of generation, reference conditioning,
editing, and continuation over a persistent visual history.  Role-aware rotary
coordinates distinguish heterogeneous visual roles, while asymmetric
clean-context prefill and fixed-budget sparse routing bound each target query's
active reads.
ConstraintSink further preserves explicit reference and source correspondences
within the same budget.  To train the sparse few-step model, privileged context
distillation transfers full-context teacher behavior through consistency
distillation and refines deployment-matched rollouts with distribution
matching.  Across the three primitive tasks and compositional trajectories,
ContextMaster improves task fulfillment and cross-shot consistency while
reaching 16 FPS.

\clearpage
\bibliographystyle{aaai2027}
\bibliography{aaai2027}

\clearpage
\appendix
\begin{center}
{\large\bfseries Appendix}
\end{center}

\section{Interactive Director}
\label{app:interactive-director}

Interactive multi-shot video creation enables users to compose generation,
reference conditioning, and editing within one evolving project.  To further
assist this process, we introduce the \emph{Interactive Director}, a
tool-augmented VLM agent inspired by Aurora~\citep{yu2026aurora}.  Given a
concise request, the Director expands it into an ordered multi-shot script and
prepares the conditions required by each operation.  It can retrieve missing
reference images from the Internet through the Serper API,
identify editing requests, and invoke OmniShotCut~\citep{wang2026omnishotcut}
to segment a source video and localize the shots to be modified.

The resulting instructions, references, and source segments are dispatched to
ContextMaster through the same generation, reference, and editing interfaces
used in our primitive tasks.  Accepted outputs are committed to the shared
visual history, allowing the Director to plan subsequent operations from the
updated project state.  Figure~\ref{fig:interactive-director} presents a
complete example of this tool-assisted creation process.

\begin{center}
    \centering
    \includegraphics[width=\columnwidth]{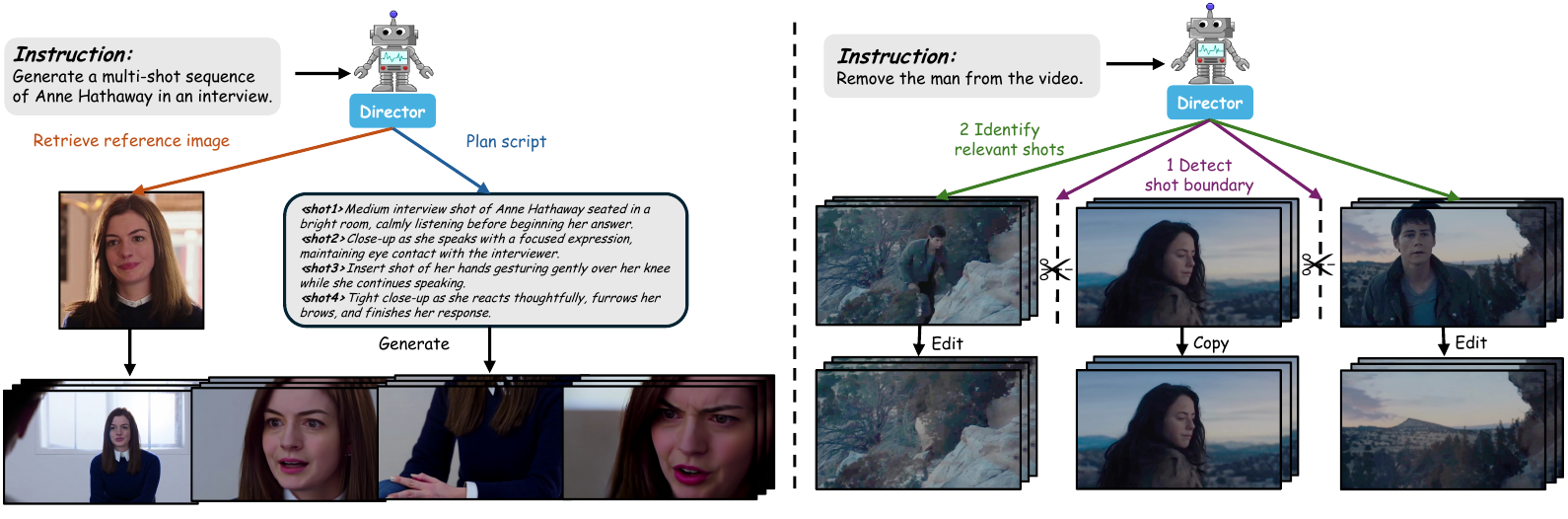}
    \captionof{figure}{Interactive multi-shot creation with the Interactive Director.
    Starting from a concise user request, the Director retrieves or prepares
    the required conditions, invokes generation and editing operations, and
    updates the accepted visual history for subsequent turns.}
    \label{fig:interactive-director}
\end{center}

\section{Implementation Details}

\subsection{Training Configuration}

Training follows the two-stage privileged context distillation strategy
described in the main paper.  We first train a dense full-context teacher with
AdamW using a learning rate of $1\times10^{-5}$, 200 warmup steps, weight decay
of 0.01, and mixed-precision training.  Privileged consistency distillation
(PCD) initializes both the sparse student and its exponential-moving-average
target from the dense teacher.

\begin{figure*}[t]
    \centering
    \captionsetup{skip=2pt}
    \includegraphics[width=\columnwidth]{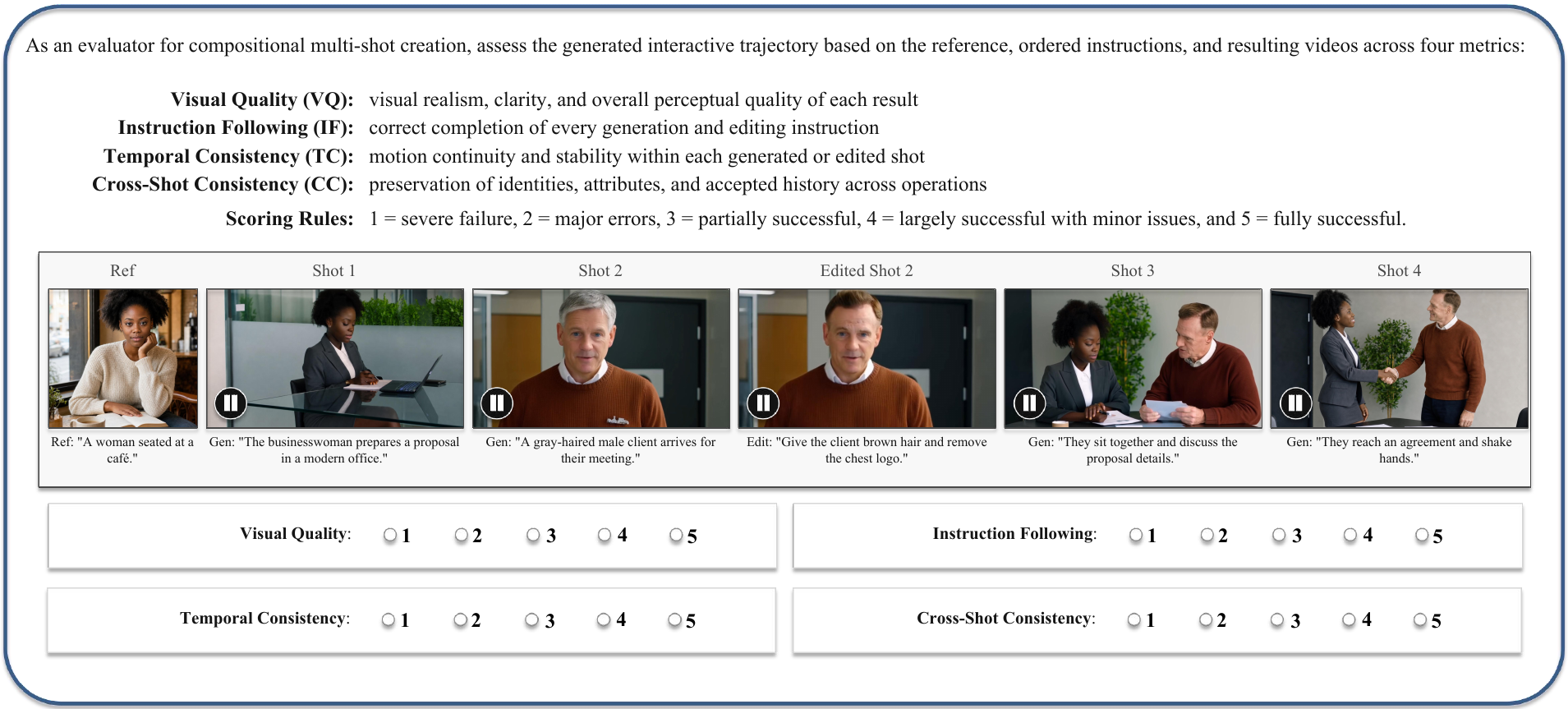}
    \caption{Interface used for the X2MV user study across four evaluation
    criteria.}
    \label{fig:user-study-interface}
\end{figure*}

PCD uses a learning rate of
$2\times10^{-6}$, an EMA decay of 0.99, a 50-level teacher noise
discretization, and the same four-step schedule used at deployment.  The
teacher applies classifier-free guidance with scale 6.0, while guidance is
absorbed into the student for single-branch inference.

Distribution matching distillation (DMD) initializes the generator from the
PCD student.  The frozen real-score model reads dense full context with
classifier-free guidance, whereas the trainable fake-score model uses the same
fixed-budget context as the generator.  The generator and fake-score model use
learning rates of $2\times10^{-6}$ and $4\times10^{-7}$, respectively, with
five fake-score updates per generator update.  All experiments are conducted on 64 NVIDIA
H200 GPUs.

\subsection{Training Data}

Teacher training begins with public HuMo~\citep{chen2025humo},
Kiwi-Edit~\citep{lin2026kiwi}, and Ditto-1M~\citep{bai2026ditto} data to
establish single-shot reference-conditioned generation and video-editing
capabilities.  We then use an internal collection of approximately one million
multi-shot videos to construct training examples for T2MV, R2MV, and V2MV.
For R2MV, we retrieve identity-matched cross-pair reference images according to
ID similarity.  For V2MV, we first segment each multi-shot video with
OmniShotCut~\citep{wang2026omnishotcut}, and then use Kling~3.0 as an expert
model to generate addition, removal, and replacement
targets.  During joint training, T2MV, R2MV, and V2MV samples are drawn at a
ratio of $4{:}3{:}3$.

\section{Additional Ablation Studies}

We ablate the active context budget to justify the deployed 6-FE setting.  All
configurations use the same checkpoint and evaluation protocol on five-shot
T2MV sequences.

\begin{table}[t]
    \centering
    \small
    \setlength{\tabcolsep}{3pt}
    \begin{tabular*}{\columnwidth}{@{\extracolsep{\fill}}ccccc@{}}
        \toprule
        Budget (FE) & AQ$\uparrow$ & Inter-Shot$\uparrow$
        & TF$\uparrow$ & FPS$\uparrow$ \\
        \midrule
        2 & 0.543 & 0.586 & 3.62 & \textbf{18.63} \\
        4 & 0.552 & 0.719 & 3.91 & \underline{17.80} \\
        6 & \underline{0.565} & \underline{0.836}
        & \underline{4.17} & 16.74 \\
        8 & \textbf{0.568} & \textbf{0.841}
        & \textbf{4.19} & 16.18 \\
        \bottomrule
    \end{tabular*}
    \caption{Context-budget ablation on five-shot T2MV.}
    \label{tab:supp-budget}
\end{table}

As shown in Table~\ref{tab:supp-budget}, increasing the budget from 2 to 6~FE
substantially improves AQ, Inter-Shot consistency, and TF.  Expanding it further
to 8~FE yields only marginal gains,
while reducing throughput by 0.56~FPS.  We therefore select 6~FE as the
quality--efficiency operating point.

\section{Evaluation Details}

\subsection{User Study}

As shown in Figure~\ref{fig:user-study-interface}, we provide the interface used
to collect human ratings across the four evaluation criteria.

\subsection{Gemini Task-Fulfillment Evaluation}

\begin{figure*}[p]
    \centering
    \includegraphics[width=0.88\textwidth]{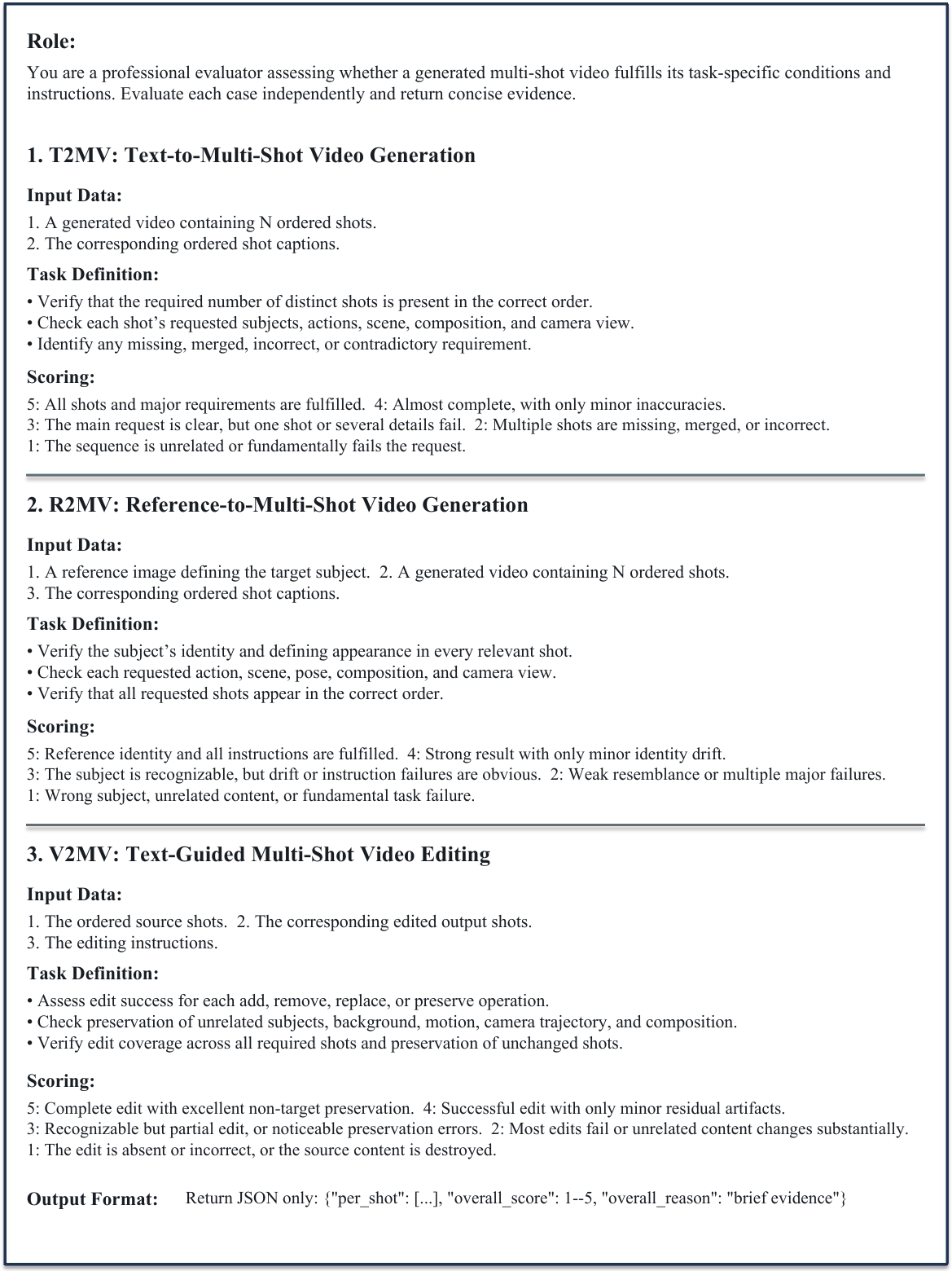}
    \caption{Prompts used by Gemini~2.5~Pro for task-fulfillment evaluation
    across T2MV, R2MV, and V2MV.}
    \label{fig:gemini-prompts}
\end{figure*}

As shown in Figure~\ref{fig:gemini-prompts}, we provide the task-specific
prompts used by Gemini~2.5~Pro to evaluate task fulfillment.

\section{Limitations}

Although fixed-budget routing bounds repeated target-to-context reads, the
context branch still performs bidirectional prefill over the full accumulated
history to construct its KV cache.
Consequently, throughput decreases gradually by approximately 0.4~FPS per
additional shot in our profiling.  Future work will explore persistent
per-shot caches and compact shot summaries.

\section{Additional Qualitative Results}

\begin{figure*}[t]
    \centering
    \includegraphics[width=\textwidth]{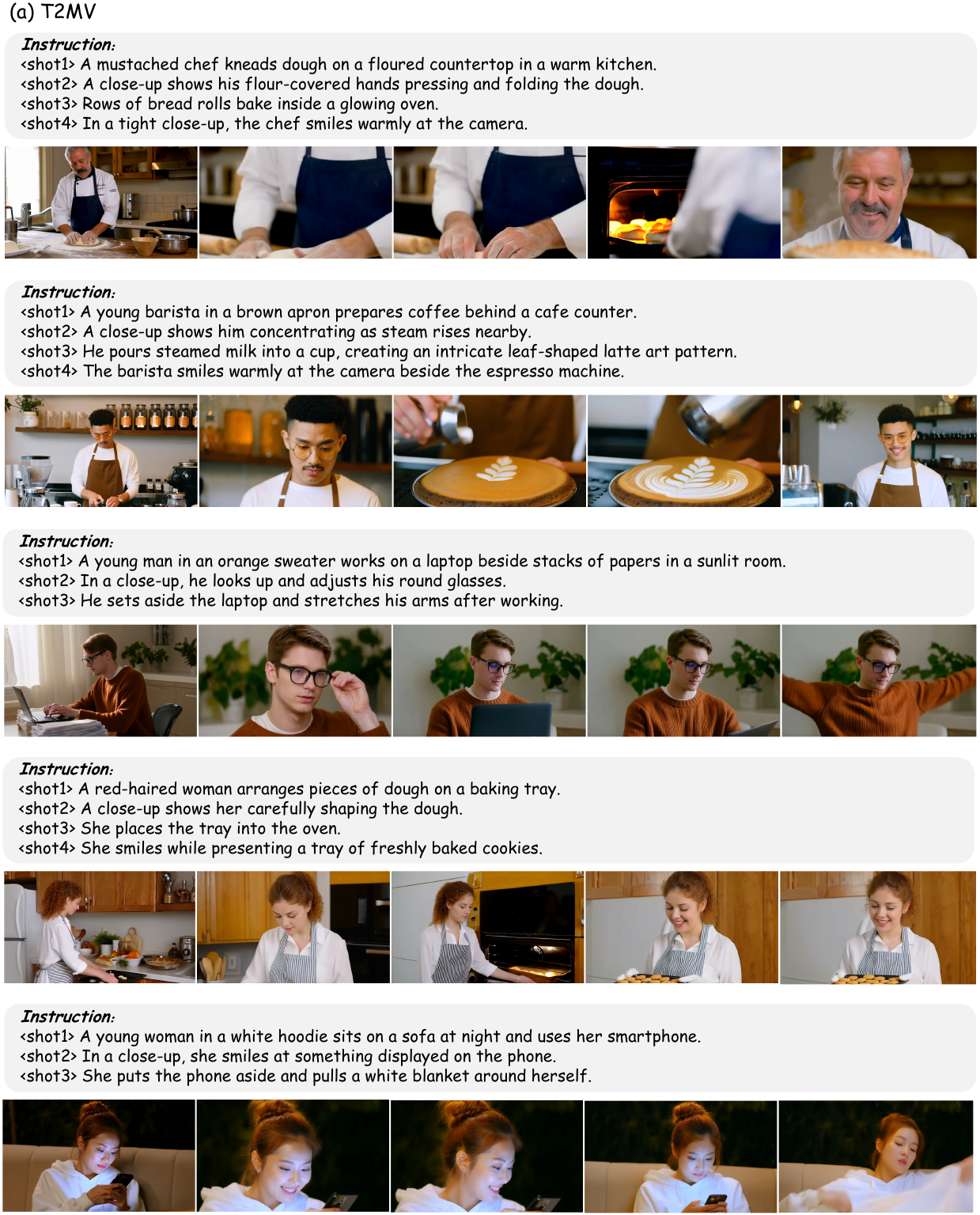}
    \caption{Additional qualitative results of ContextMaster on T2MV.}
    \label{fig:supp-t2mv}
\end{figure*}

\begin{figure*}[t]
    \centering
    \includegraphics[width=\textwidth]{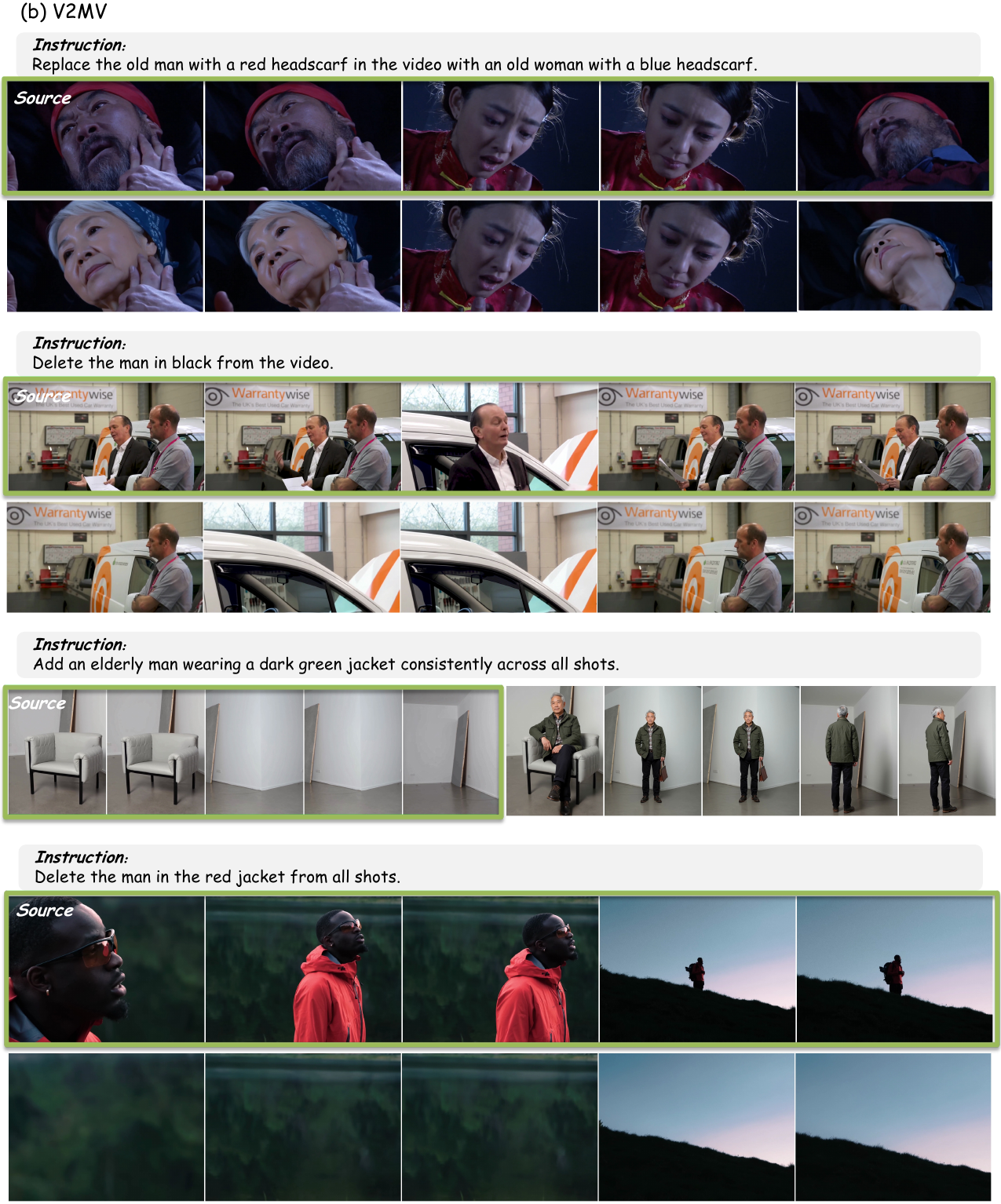}
    \caption{Additional qualitative results of ContextMaster on V2MV.}
    \label{fig:supp-v2mv}
\end{figure*}

\begin{figure*}[t]
    \centering
    \includegraphics[width=\textwidth]{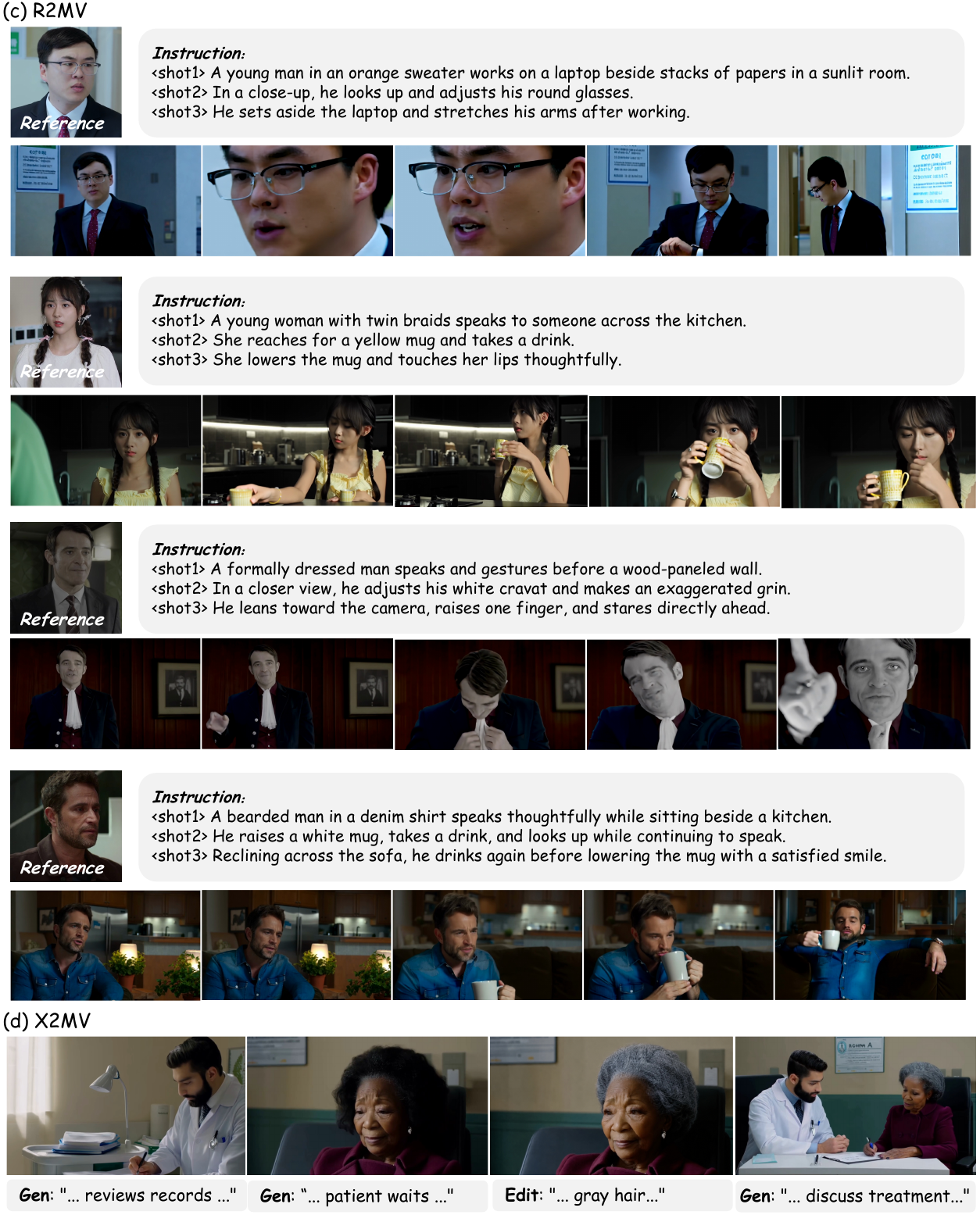}
    \caption{Additional qualitative results of ContextMaster on R2MV and X2MV.}
    \label{fig:supp-r2mv-x2mv}
\end{figure*}

As shown in Figures~\ref{fig:supp-t2mv}--\ref{fig:supp-r2mv-x2mv}, we provide
additional qualitative results of ContextMaster on T2MV, V2MV, R2MV, and X2MV.

\clearpage

\end{document}